\documentclass{article}

\usepackage{PRIMEarxiv}
\usepackage[utf8]{inputenc} 
\usepackage[T1]{fontenc}    
\usepackage{hyperref}       
\usepackage{url}            
\usepackage{booktabs}       
\usepackage{amsfonts}
\usepackage{amsmath,amssymb}
\usepackage{nicefrac}       
\usepackage{microtype}      
\usepackage{lipsum}
\usepackage{fancyhdr}       
\usepackage{graphicx}       
\graphicspath{{figures/}}
\usepackage{fancyhdr}
\usepackage{algorithm}
\usepackage{algpseudocode}
\usepackage{subcaption}

\title{Enhancing AI Diagnostics: Autonomous Lesion Masking via Semi-Supervised Deep Learning
}

\author{
  Ting-Ruen Wei, Michel Hell, Dang Bich Thuy Le, Yuling Yan\thanks{Corresponding Author}\\
  Santa Clara University \\
  Santa Clara, CA\\
  \texttt{\{twei2, mhell, dle, yyan1\}@scu.edu} \\
   \And
  Aren Vierra, Ran Pang, Mahesh Patel, Young Kang \\
  Santa Clara Valley Medical Center \\
  San Jose, CA\\
}

\begin{document}
\maketitle

\begin{abstract}
This study presents an unsupervised domain adaptation method aimed at autonomously generating image masks outlining regions of interest (ROIs) for differentiating breast lesions in breast ultrasound (US) imaging. Our semi-supervised learning approach utilizes a primitive model trained on a small public breast US dataset with true annotations. This model is then iteratively refined for the domain adaptation task, generating pseudo-masks for our private, unannotated breast US dataset. The dataset, twice the size of the public one, exhibits considerable variability in image acquisition perspectives and demographic representation, posing a domain-shift challenge. Unlike typical domain adversarial training, we employ downstream classification outcomes as a benchmark to guide the updating of pseudo-masks in subsequent iterations. We found the classification precision to be highly correlated with the completeness of the generated ROIs, which promotes the explainability of the deep learning classification model. Preliminary findings demonstrate the efficacy and reliability of this approach in streamlining the ROI annotation process, thereby enhancing the classification and localization of breast lesions for more precise and interpretable diagnoses.
\end{abstract}


\section{Introduction}
Classification tasks are at the core of deep learning, exemplifying the adaptability and scalability of deep neural network architecture and components. A noteworthy development is the emergence of high-performing convolutional neural networks (CNNs), bolstered by the adoption of transfer learning, a practice that leverages the pre-trained models and fine-tunes them for domain-specific tasks, particularly those handling image data. This approach offers the dual benefits of reduced training time and enhanced model efficacy. However, it is important to recognize that, while transfer learning has demonstrated its effectiveness, the task of adapting pre-trained models from a generic landmark dataset to the intricacies of specific medical domains is not always a straightforward process. Model generalization, denoting the capability of a model to extend its knowledge beyond the training data, presents a formidable obstacle. In particular, variances in data distribution and image acquisition protocols, as well as the diverse perspectives they encompass pose substantial challenges in achieving robust generalization while maintaining diagnostic accuracy. This ongoing challenge calls for continuous research and innovation.

In the realm of AI-assisted medical diagnosis, the prevailing approaches, such as CNN models and transfer learning, are fundamentally rooted in supervised learning. A critical stage in this process involves labeling to establish ground truth and annotating to enhance classification or detection outcomes. However, this process is labor-intensive and often becomes a bottleneck in model development. While the image-level labeling is straightforward, more detailed annotation tasks, such as drawing bounding boxes or masks, pose significant challenges. The limited availability of high-quality annotated data and the potential for inter-observer variability in annotations both substantially impact model performance. Another limitation specific to medical datasets is patient privacy; as opposed to other domains, the Health Insurance Portability and Accountability Act and Institutional Review Boards safeguard the distribution of protected health information. This regulation significantly limits the availability of public datasets and makes transfer learning within the medical domain a more challenging task. However, addressing this challenge is notably rewarding. The ideal scenario is to transfer the representation learned from large datasets within the same domain. Yet, acquiring a substantial dataset specific to that domain can be challenging. As a result, domain adaptation within the medical field becomes particularly rewarding. This approach holds the potential to assimilate knowledge from existing datasets and expand their utility to less developed domains. While there is existing research on unsupervised domain adaptation, it primarily focuses on virtual-to-realistic applications, which bear little relevance to the intricacies of medical applications handling the complexities of the human body. A common domain shift in medical imaging is due to the use of different modalities leading to different intensity distribution or image structures. By implementing domain adaptation from the most similar existing dataset and enabling the transfer learning thereafter, we open the gateway for more effective and efficient generalization across diverse applications.

In this study, we take a proactive step to address challenges posed by data scarcity and annotation complexities in medical image analysis. Our proposed unsupervised domain adaptation approach is designed to autonomously generate and iteratively refine image masks outlining breast lesion ROIs for the classification of benign from malignant breast masses. The method dynamically adjusts the generated pseudo-masks in response to evolving classification outcomes, effectively mitigating large variations within the different datasets. Our approach leverages a small set of meticulously annotated data (with image masks), combined with a large private pool of dataset without image masks for model training and enhancement. Through this iterative self-training process, we optimize performance for the downstream classification tasks, and the generated ROIs can be utilized to interpret the classification or assist human readers for more accurate diagnoses \cite{shen2021artificial}. Notably, only simple image-level labels (benign or malignant) are required to infer the presence of breast cancer in the final classification. This innovative approach represents a promising solution to the challenges of data scarcity  and annotation complexity, providing a robust foundation for more effective medical diagnoses. 

Our key contribution is the introduction of a novel unsupervised domain adaptation framework tailored for medical image segmentation. This framework was strategically designed to promote the explainability of the downstream model, exemplified in our study on breast cancer detection using US images. Demonstrating the feasibility of our approach, we applied it to adapt from a public to a private US breast image dataset and our results showcased the effectiveness of our framework and its potential for broader applications within the medical domain. By addressing the challenge of domain shifts, our approach may open avenues for more robust and generalized solutions in medical image analysis, ultimately contributing to advancements in AI-assisted diagnosis and treatment.

The field of semantic segmentation using deep learning has been widely studied. Due to the practical limitation of fully supervised methods, we focus on semi-supervised and unsupervised techniques.

\noindent\textbf{Semi-supervised Semantic Segmentation.} Many existing works revolve around the concepts such as consistency training, contrastive learning, and self-training with pseudo-maps. Consistency training aims to minimize the difference between augmented images under perturbations. Liu et al. \cite{liu2022perturbed} utilized a confidence-weighted cross-entropy loss to improve generalization within a teacher-student model \cite{tarvainen2018mean}. Contrastive learning draws positive pairs closer and moves them away from negative pairs. PC$^2$Seg \cite{zhong2021pixel} leveraged pixel-wise contrastive loss in the feature space. Self-training, incorporating pseudo-maps into the labelled data for iterative training, is used in our approach. Similarly, ST++ \cite{yang2022st++} focused on the selection of more reliable unlabeled data for re-training, while AEL \cite{hu2021semi} balances for the under-performing category, ELN \cite{kwon2022semi} identifies pixel-wise error locations, and U$^2$PL \cite{wang2022semi} separates reliable from unreliable pixels category-wise by their entropy to form negative samples. Other studies direct their attention to activation maps \cite{Lee_2019_CVPR, wei2018revisiting, lee2021anti} and generative adversarial networks \cite{hung2018adversarial, souly2017semi, Li_2021_CVPR}. however, studies in this area operate within the scope of a single dataset and cannot easily adapt to multiple sources of datasets. A few exceptions such as the work of Li et al. \cite{Li_2021_CVPR} and Kalluri et al. \cite{kalluri2019universal} exists but with different limitations. In particular, Li et al. use GANs and the approach is unsuitable for real-time applications critical for medical diagnosis. The work of Kalluri et al. mandates a portion of labelled data in input datasets, imposing a further constraint on the domain adaptation task in medical datasets. Some works in semi-supervised semantic segmentation \cite{wang2022cnn, luo2021semi, bortsova2019semi, chen2019multi, nie2018asdnet, hu2021semi_medical} directly tackle medical datasets, and Zhou et al. \cite{Zhou_2019_CVPR} further combine the segmentation and classification task, similar to our purpose, but they do not allow domain adaptation.

\noindent\textbf{Unsupervised Domain Adaptation}
Existing approaches accomplish the adaptation by alignment \cite{He_2021_CVPR, zhang2019category, Yang_2020_CVPR, Li_2019_CVPR} or adversarial learning \cite{Choi_2019_ICCV, Pan_2020_CVPR, huang2020contextualrelation, Vu_2019_CVPR, Du_2019_ICCV}. The work of Yang and Soatto \cite{Yang_2020_CVPR} align the source and target domains via Fourier transform, and Choi et al. \cite{Choi_2019_ICCV} use GANs in self-ensembling. Other works like GUDA \cite{Guizilini_2021_ICCV}, MADAN \cite{zhao2019multisource}, MaxSquare \cite{Chen_2019_ICCV}, and Yang et al. \cite{Yang_2021_WACV} adopt various strategies, such as monocular depth estimation, allowing for multiple source datasets, implementation of maximum squares loss to balance towards difficult samples, and application of cross-attention for context dependency. Similar to our approach, CBST \cite{zou2018unsupervised}, IAST \cite{mei2020instance}, and Subhani and Ali \cite{subhani2020learning} leverage self-training. Of particular relevance to the medical domain, UBNA \cite{klingner2021unsupervised} and SFDA \cite{liu2021sourcefree} have achieved domain adaptation without requiring the access of source images, which benefits the patient privacy. However, all studies in this area focus on the virtual-to-realistic domain adaptation from GTA-5 \cite{richter2016playing} or SYNTHIA \cite{ros2016synthia} to Cityscapes \cite{cordts2016cityscapes}. To expand the scope of unsupervised domain adaptation beyond these specific cases, we propose a versatile framework for datasets with an unknown domain shift and this is exemplified through demonstration with two US breast image datasets. In this way, we illustrate the adaptability of our approach to various applications, especially in the medical field where datasets are small and annotations are limited. Notably, Perone et al. \cite{perone2019unsupervised} have effectively utilized self-ensembling on public Magnetic Resonance Imaging (MRI) datasets for segmentation tasks. The outcome of medical image segmentation significantly impacts the classification of pathology, as recognized by numerous researchers in the field \cite{Zhou_2019_CVPR}. In this study, we prioritize the classification task of differentiating benign from malignant breast masses and utilize it as a benchmark to guide the iterative refinement process for generating lesion ROIs in our domain adaptation task.

\section{MATERIALS AND METHODS }

\begin{figure}[t]
  \centering
   \includegraphics[width=\linewidth]{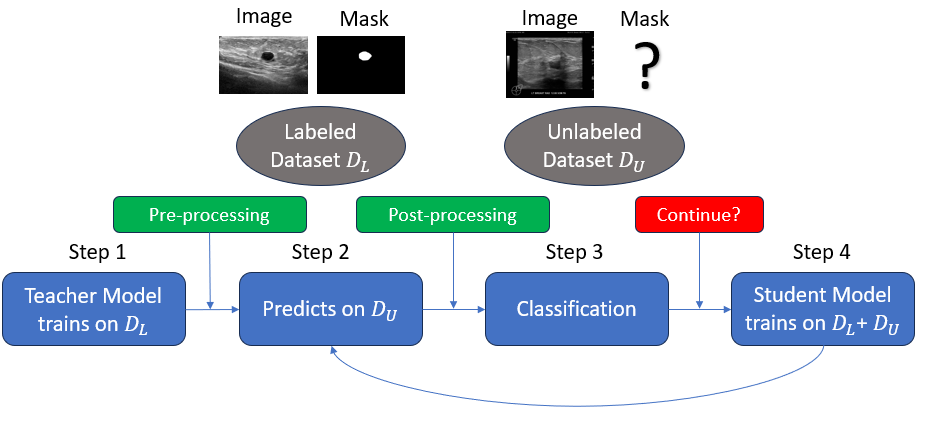}
   \caption{Overview of our proposed approach. We train Teacher Model on the labeled dataset $D_L$ and iteratively repeat the following: predict on the unlabeled dataset $D_U$ for pseudo-masks, perform downstream classification, and train Student Model on the combined dataset. Self-training terminates when classification performance is satisfactory.}
   \label{fig:flowchart}
\end{figure}

The proposed new learning approach involves training an initial "teacher" model on a labelled dataset $D_L$, followed by the pseudo-label generation, model retraining and iterative refinement, as shown in Figure \ref{fig:flowchart}. The corresponding pseudo-code is presented in Algorithm \ref{alg}.

\begin{figure}
  \centering
\includegraphics[width=0.5\linewidth]{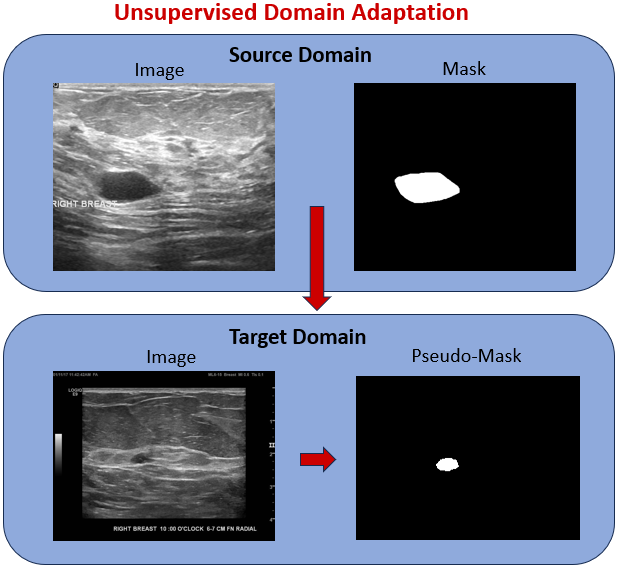}
   \caption{Our proposed approach for unsupervised domain adaptation. Given the domain shift between source and target domains, we propose a framework to generate and refine the pseudo-masks for the target domain, which are then used for downstream classification. (The pseudo-mask presented here is obtained after iterations of self-training.)}
   \label{fig:intro}
\end{figure}

An illustration of the proposed unsupervised domain adaptation method is presented in Figure \ref{fig:intro}. The top row displays an example dataset from the source domain, including both a raw image (left) and the corresponding mask (right). In the bottom row, a raw image (left) and the predicted pseudo-mask (right) are depicted. The pseudo-masks are generated and utilized for the downstream classification task.

\textbf{Step 1: Initial Teacher Model}. The approach starts with an initial segmentation model, referred to as the "teacher model." This teacher model is trained on a set of reference data which is a public dataset that includes both images and corresponding annotations, or reference masks. The teacher model learns to segment the regions of interest (ROI) in the images based on the labeled data. This minimizes the cross-entropy loss defined as:

\begin{equation}
  -\frac{1}{N} \sum_{i=1}^{N} \sum_{z=1}^{Z} y_{iz} log(\hat{y}_{iz})
  \label{eq:CE}
\end{equation}

\noindent where $N$ is the total number of pixels, $y_{iz}$ is whether pixel $i$ belongs to class z, and $\hat{y}_{iz}$ is the probability of pixel $i$ belonging to class z.

\begin{algorithm}[!t]
\caption{Our Proposed Approach}
\label{alg}
\begin{algorithmic}
\Require Labeled Dataset $D_L$ with (image, mask) as ($x_L^i$, $y_L^i$), Unlabeled Dataset $D_U$ with (image, label) as ($x_U^i$, $l_U^i$), Segmentation model $M_S$, Classification Model $M_C$
\While{TRUE}
    
    Train $M_S$ on $D_L$
    
    Predict all $x_U^i$ with $M_S$ for $\hat{y}_U^i$

    Segment $x_U^i$ with $\hat{y}_U^i$ for $\Bar{x}_U^i$

    Classify $\Bar{x}_U^i$ for $l_U^i$ with $M_C$

    \If{Satisfactory}
    
            break
    \EndIf

    Copy some ($x_U^i$, $\hat{y}_U^i$) to $D_L$ with data filtering
\EndWhile
\end{algorithmic}
\end{algorithm}

\textbf{Step 2: Pseudo-Label Generation}. Once the teacher model is trained, we utilize it to generate new segmented masks for a batch of unlabeled data $D_U$ which is a private dataset. This batch consists of a large volume of images that come from a different source. The teacher model's predictions on these unlabeled images serve as pseudo-labels, acting as if they were true annotations. These pseudo-labels represent the model's best guess at segmenting the objects in the unlabeled data.

\textbf{Step 3: Downstream Classification}. We segment the images with the pseudo-masks and perform classification on the cropped images. This step is optional in the first iteration as the pseudo-masks are heavily noisy due to the nature of unsupervised domain adaptation. We use the performance of the validation set in this step to determine if the entire procedure should repeat for another iteration.

\textbf{Step 4: Training Student Model}. If we continue for another iteration, we initialize the student model with the teacher model and train it on a combination of the source domain (true-masks) $D_L$ and the newly created pseudo-labeled data (pseudo-masks) $D_U$. This training process helps the model improve its segmentation capabilities by learning from a larger and more diverse dataset that includes both labeled and pseudo-labeled examples.

\textbf{Iterative Refinement}. The student model becomes the teacher model and now updates the pseudo-masks through inference. This process repeats steps 2 to 4 until termination by satisfactory classification performance in step 3. This iterative cycle can be repeated multiple times to further enhance the model's performance. The unsupervised domain adaptation is illustrated in Figure \ref{fig:intro}. The key idea behind this approach is to leverage accessible labeled data and progressively refine the segmentation model by incorporating predictions on unlabeled data. The self-training process aims to improve the model's generalization and segmentation effectiveness by learning from its own predictions in an iterative manner. It is particularly useful when labeled data is scarce, yet large volumes of unlabeled data are available from different sources. As opposed to reaching high intersection-over-union scores, which requires the time-consuming annotation maps and many datasets do not have, we segment for the region of interest that the downstream classification task can benefit from as well. In this way, we can properly leverage the available data and avoid the domain shift. The final ROIs can serve to interpret the classification model.

\textbf{Pre-processing}. The initial pre-processing layers in the Teacher segmentation model operates on a pixel-wise basis. However, the source and target domains can have an entirely different visual structure, as encountered in our case. The pre-processing can align the two domains in a non-parametric way. For our dataset, we adopt the entropy filter as follows. The notion of entropy is intricately linked to the level of disorder within a physical system, and it is redefined as a measure of the amount of information or uncertainty present in a source \cite{ShannonEntropy}. If the source is a 2D array of information, like an image, the entropy redefined by Shannon in \cite{ShannonEntropy} is given by:

\begin{equation} \label{eq:entro1}
    H(X) = - \sum_{i = 1}^{n} p(x_i) \log_b p(x_i)
\end{equation}

\noindent where $Pr [X = x_i] = p(x_i)$ is the probability mass distribution of the source. In this way, Equation~\ref{eq:entro1} can be used to estimate the global entropy of an image characterized by its histogram:

\begin{equation}
    H(I) = - \sum_{i = 1}^{N} \mbox{hist}_{\mbox{norm}}(L_i) \log \left(\mbox{hist}_{\mbox{norm}}(L_i) \right) 
\end{equation}

\noindent where $L_i$ represents the $N$ intensity levels of the $m \times n$ image $I(x, y)$ and $\mbox{hist}_{\mbox{norm}}(L_i)$ is the histogram properly normalized to fit a probability distribution function:

\begin{equation}
    \sum_{i = 1}^{N} \mbox{hist}_{\mbox{norm}}(L_i) = 1
\end{equation}

In this study, entropy is used to analyze the complexity of the private ultrasound images to find the most informative portion of the image and then differentiate the effective exam area from the surrounding information, such as texts, rules and black areas. The objective is to enhance the similarity of general image characteristics between the private and public databases, promoting a smoother and more effective domain shift.

\begin{figure}
  \centering
    \begin{subfigure}{0.18\linewidth}
    \includegraphics[width=\linewidth]{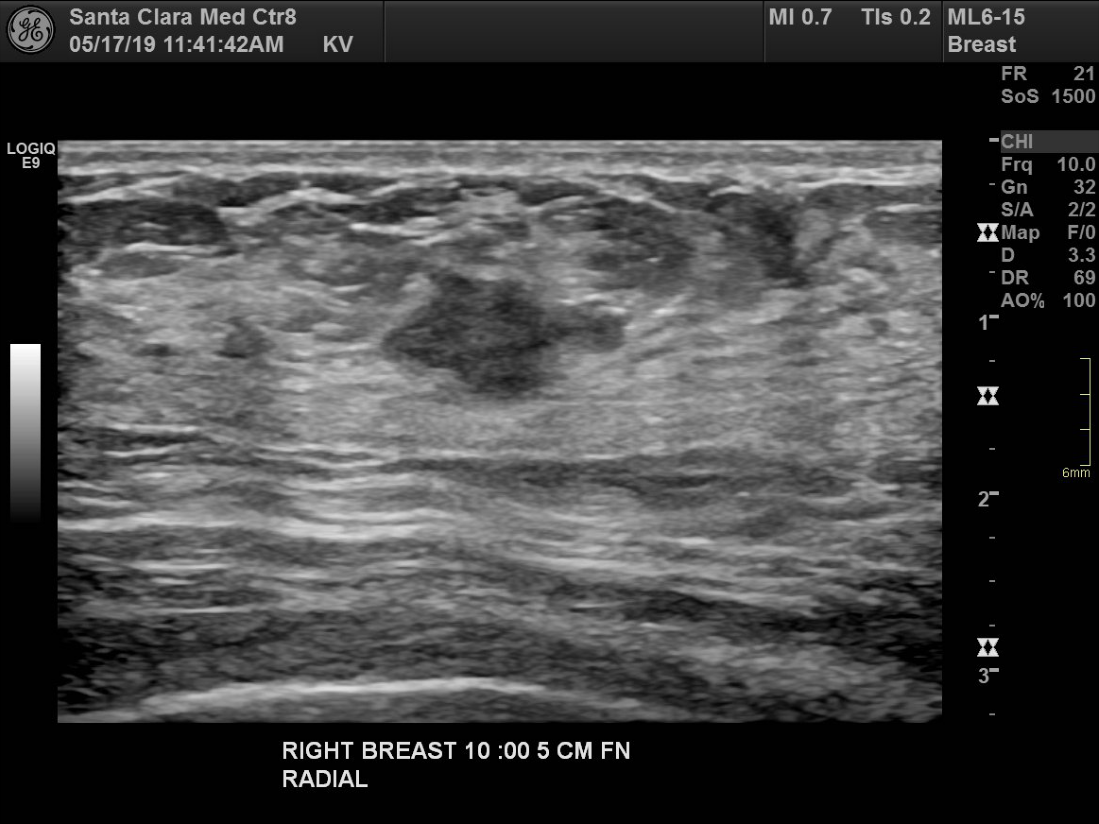}
    \caption{}
    \label{fig:pol-a}
  \end{subfigure}
  \hfill
  \begin{subfigure}{0.18\linewidth}
    \includegraphics[width=\linewidth]{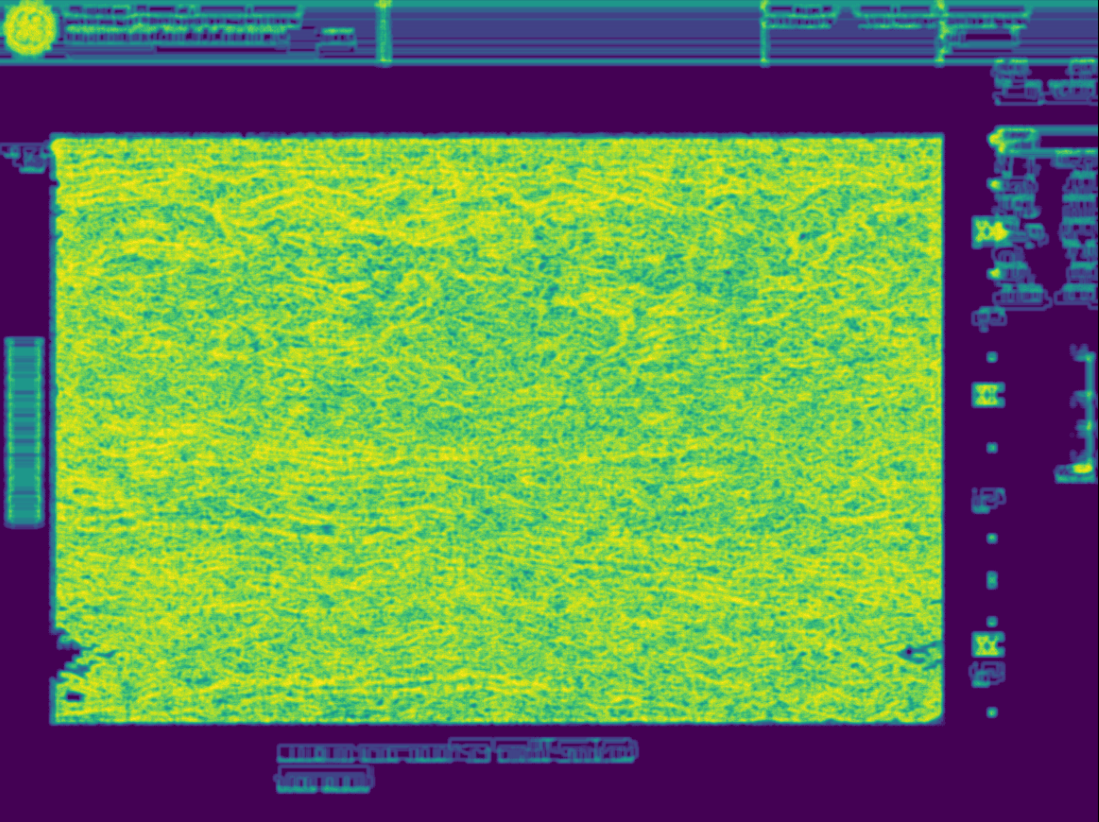}
    \caption{}
    \label{fig:pol-b}
  \end{subfigure}
    \hfill
  \begin{subfigure}{0.18\linewidth}
    \includegraphics[width=\linewidth]{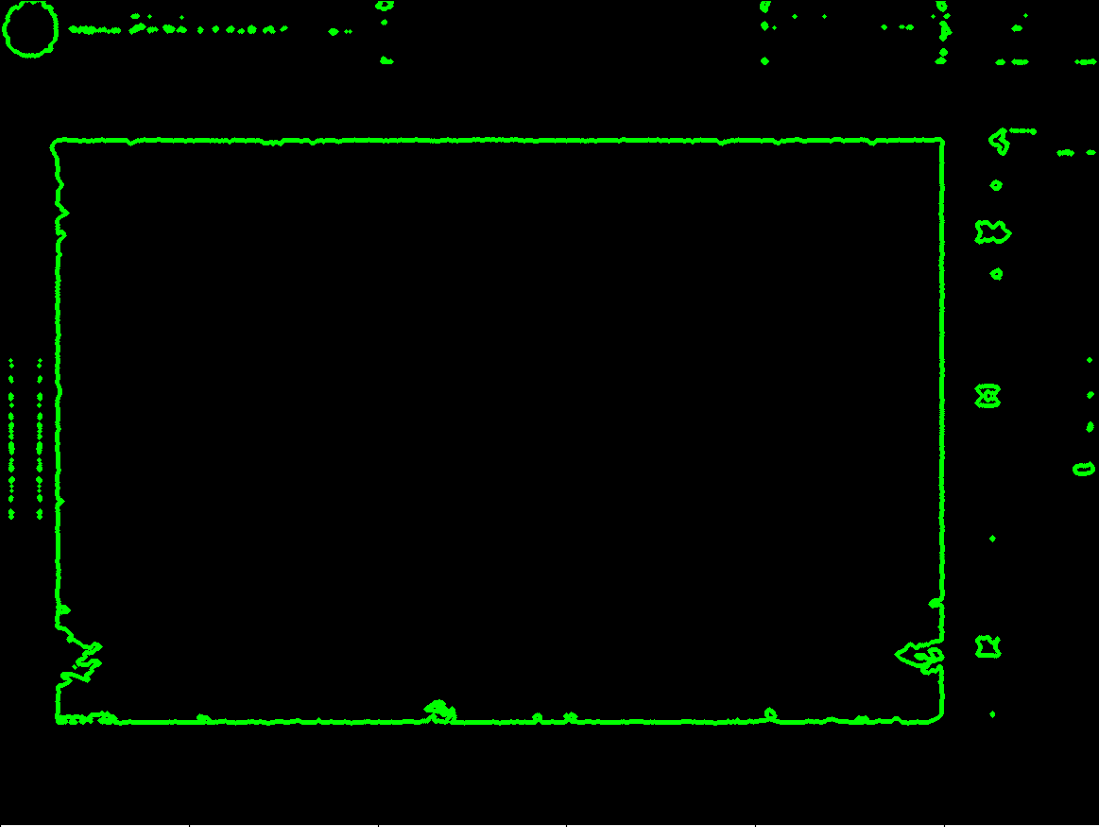}
    \caption{}
    \label{fig:pol-c}
  \end{subfigure}
  \hfill
  \begin{subfigure}{0.18\linewidth}
    \includegraphics[width=\linewidth]{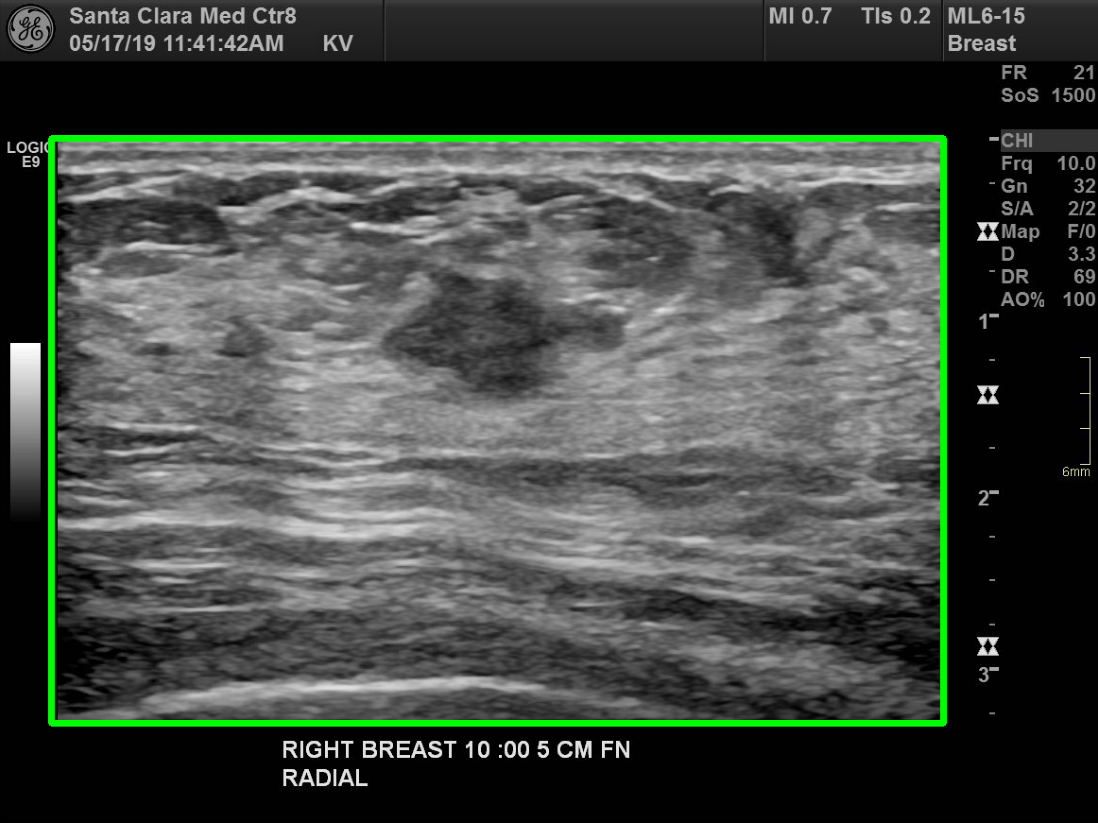}
    \caption{}
    \label{fig:convex-a}
  \end{subfigure}
  \hfill
  \begin{subfigure}{0.18\linewidth}
    \includegraphics[width=\linewidth]{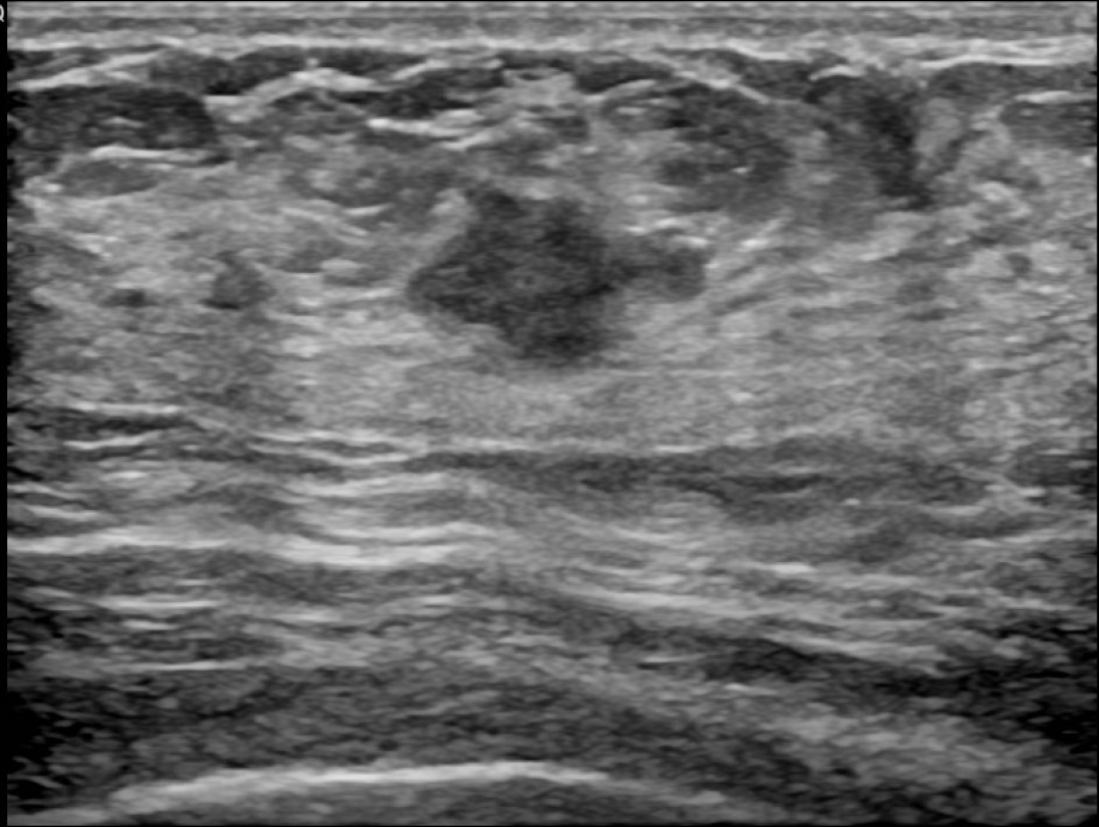}
    \caption{}
    \label{fig:convex-b}
  \end{subfigure}
  
  \caption{Cropping an image with entropy filter. (a) Original Image (b) Entropy filtered image (c) Contours of Entropy filtered image (d) Identification of exam area (e) Final cropped image}
  \label{fig:entropy}
\end{figure}

Figure~\ref{fig:entropy} illustrates the pre-processing methodology employed in this study. Initially, the entropy filter is implemented on the original private image (Fig.~\ref{fig:entropy}(a)), yielding the filtered image presented in Fig.~\ref{fig:entropy}(b). Subsequently, the contours of this filtered image are discerned (Fig.~\ref{fig:entropy}(c)), and the examination area is defined as the region encompassed by the largest contour as can be seen in Fig.~\ref{fig:entropy}(d). Finally, the image is cropped to retain solely the identified exam area resulting in the image presented in Fig.~\ref{fig:entropy}(e).

\begin{figure*}
  \centering
    \begin{subfigure}{0.13\linewidth}
    \includegraphics[width=\linewidth]{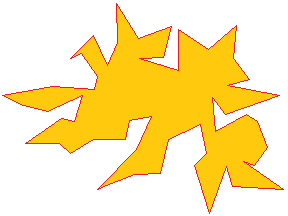}
    \caption{}
    \label{fig:pol-a}
  \end{subfigure}
  \hfill
  \begin{subfigure}{0.13\linewidth}
    \includegraphics[width=\linewidth]{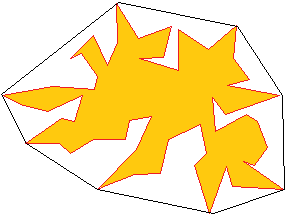}
    \caption{}
    \label{fig:pol-b}
  \end{subfigure}
    \hfill
  \begin{subfigure}{0.13\linewidth}
    \includegraphics[width=\linewidth]{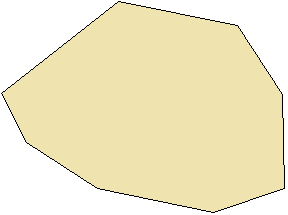}
    \caption{}
    \label{fig:pol-c}
  \end{subfigure}
  \hfill
  \begin{subfigure}{0.13\linewidth}
    \includegraphics[width=\linewidth]{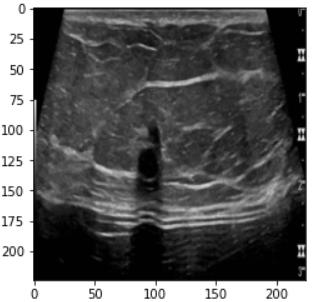}
    \caption{}
    \label{fig:convex-a}
  \end{subfigure}
  \hfill
  \begin{subfigure}{0.13\linewidth}
    \includegraphics[width=\linewidth]{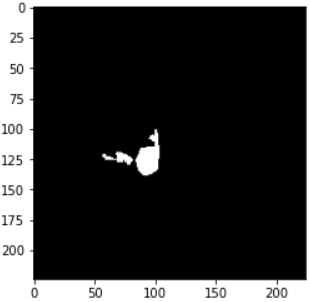}
    \caption{}
    \label{fig:convex-b}
  \end{subfigure}
    \hfill
  \begin{subfigure}{0.13\linewidth}
    \includegraphics[width=\linewidth]{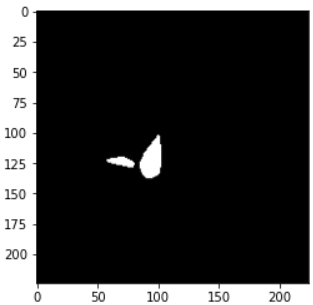}
    \caption{}
    \label{fig:convex-c}
  \end{subfigure}
  \hfill
  \begin{subfigure}{0.13\linewidth}
    \includegraphics[width=\linewidth]{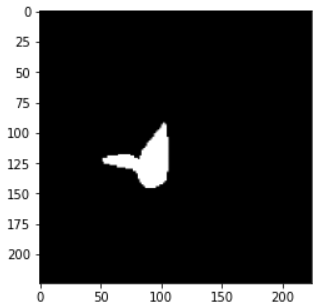}
    \caption{}
    \label{fig:convex-d}
  \end{subfigure}
  
  \caption{Convex hull operation. (a) Non-convex area (b) Convex hull process (c) Final Convex hull (d) Input Image (e) Pseudo-mask (f) Convex hull of pseudo-mask (g) 1.5x scaled convex hull.}
  \label{fig:convex2}
\end{figure*}

\textbf{Post-processing}. Since noise is inherent in the pseudo-masks during domain adaptation, we apply post-processing to refine them. For our dataset, we employ hole filling and the convex hull on the pseudo-masks before classification. We utilize the SciPy module to fill holes in the enclosed regions, leveraging the prior knowledge that masks are fully enclosed. This enhances the learning of segmentation maps in the next iteration of self-training. As some pseudo-masks may be blank due to the domain shift, providing no information for the next iteration of self-training, we enforce an additional logic in the pseudo-mask update: if the new pseudo-mask is blank, we do not replace the previous one. Although large masks can be replaced by much smaller ones, the convex hull described below can enlarge it to include peripheral information. Since the main task is downstream classification and we want to leverage the segmentation maps in the source domain without being compromised with domain shift, we obtain the convex hull to include neighboring pixels for regularization. In cases where tiny masks are generated due to the domain shift, we scale the convex hull to a larger region to compensate. A trained segmentation model predicts the cropped image (Fig.~\ref{fig:convex2}(d)) to have a mask (Fig.~\ref{fig:convex2}(e)). We obtain its convex hull (Fig.~\ref{fig:convex2}(f)) and scale it by 150\% (Fig.~\ref{fig:convex2}(g). Specifically, we identify all contours with an area greater than zero and combine their convex hull with the OR operation. Optionally, larger masks can be scaled down to reduce noise. The convex hull denotes the smallest convex polygon that overlaps with all points in the given region and is described as the following:

Given a set of $n$ points $P = { p_1, \ldots , p_n } \subset \mathbb{R}^q$, $n \in \mathbb{N}$, the convex hull $C$ is defined as the intersection of all convex sets containing $P$ and it is then given by the expression:

\begin{equation} \label{eq:convex}
    C \equiv \left\{  \sum_{i = 1}^{n} \lambda_i p_i : \lambda_i \geq 0 ~\forall i = 1, \ldots, n,  \sum_{i = 1}^{n} \lambda_i = 1\right\} 
\end{equation}

\begin{figure}
  \centering
   \includegraphics[width=0.5\linewidth]{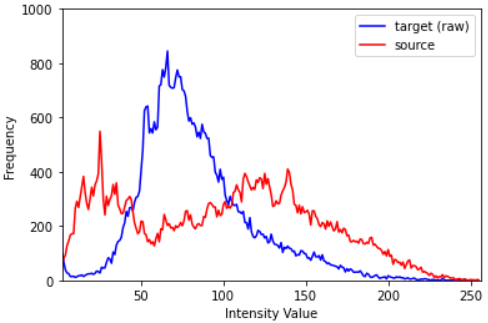}
   \caption{Intensity distribution of source and pre-processed target images.}
   \label{fig:intensity}
\end{figure}

In practical terms, the application of the Equation~\ref{eq:convex} results in the generation of the smallest polygon that encompasses all points $p_i$, $i = 1, \ldots, n$ in $P$ while ensuring the absence of inwardly bent corners in the resulting polygon.

In particular, for binary images like the masks produced by the teacher model, the implementation of Equation~\ref{eq:convex}  results in the generation of the smallest polygon that encompasses the entire mask area while ensuring the absence of inwardly bent corners in the resulting mask producing in this way more suitable masks for the domain shift process. An example of this process is shown in Figure~\ref{fig:convex2}.

\textbf{Data Filtering}. Before the labeled and unlabeled datasets combine for Step 4, data filtering \cite{xie2020selftraining} is applied to select more confident pseudo-masks. This process depends on prior knowledge of the dataset, but in general, tiny pseudo-masks do not contribute significantly to the next training iteration, and we have greater confidence that larger masks will encompass the true ROI. Based on observing the source domain, we think that pseudo-masks with one connected component large enough entails high confidence for our specific dataset. Therefore, we filter for bigger and connected pseudo-masks and add them into the training set for the next iteration. The hole filling operation mentioned earlier assists with this process by producing more connected pseudo-masks and enhancing the segmentation model.

\textbf{Datasets}
We use a public dataset on US breast images \cite{al2020dataset} as our source domain and a private dataset on US breast images provided by Hospital 1 as the target domain. The source domain is a fully labelled dataset with mask annotations, comprising 665 pairs of (image, binary mask). The target domain comprises 1182 images, including 796 benign and 386 malignant masses, serving as the basis for downstream classification. The images are acquired in two perspectives, radial and anti-radial views. As shown in Figure \ref{fig:intensity}, the two datasets exhibit distinct intensity distributions, along with noticeable differences in visual structures, as evident in Figure \ref{fig:intro}. For classification testing, 10\% of the target domain is allocated for classification and 15\% is assigned to validation set for hyperparameter tuning.

\textbf{Model Training}
For semantic segmentation, we use UNet \cite{ronneberger2015unet} with an input shape of 224 by 224. The model downsamples with max pooling and convolutional layers to an intermediate shape of 14 by 14 by 256 and subsequently upsamples to the original input shape. The output is a semantic segmentation map with each pixel representing the probability of being activated in the binary mask. The model is trained for 100 epochs with early stopping in the first iteration and fine-tuned for 10 epochs in the subsequent iterations. As for the downstream classification, we follow the common transfer learning approach by utilizing a pre-trained DenseNet169 \cite{huang2018densely} which is known for encouraging feature reuse and reducing the total number of parameters. 

\section{RESULTS}

\begin{figure}[b]
  \centering
   \includegraphics[width=0.5\linewidth]{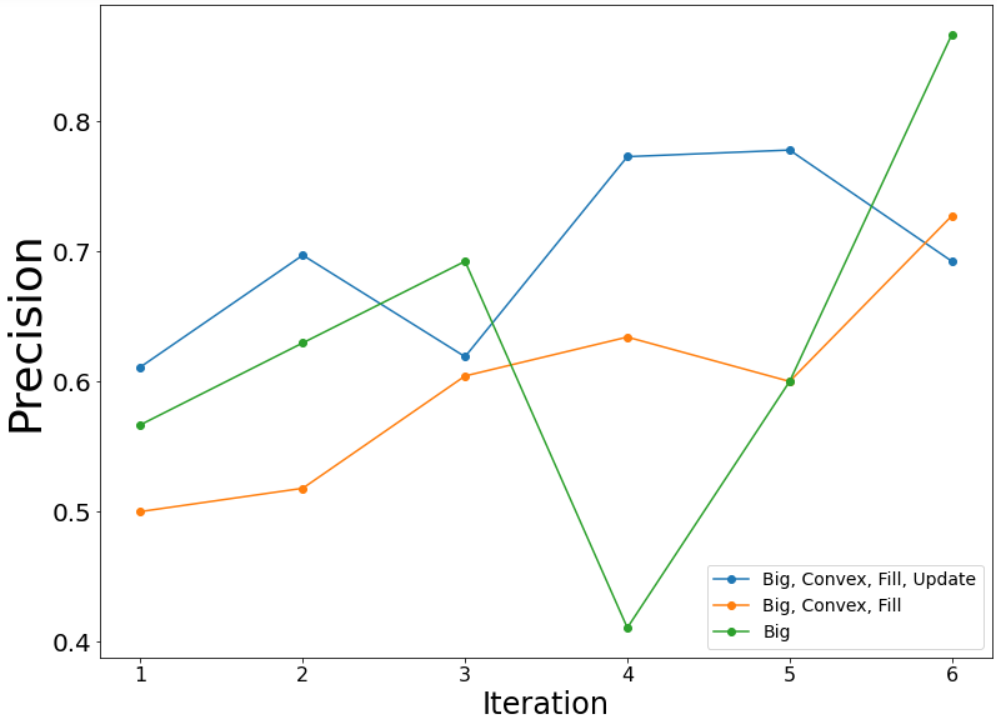}
   \caption{Precision curve across iterations. We observe a correlation between performance and mask completeness in Figure \ref{fig:results}}
   \label{fig:precision}
\end{figure}

With the pseudo-masks generated by the segmentation model, we segment the pre-processed images and perform the downstream classification. Since irregular contour is one feature that differentiates malignant from benign masses, we obtain the smallest rectangle that fully covers the pseudo-mask output and use it for the final classification. 

Figure \ref{fig:precision} shows the precision curve for the mixture of techniques we employed over 6 iterations of self-training. The vanilla model (colored in brown) is among the bottom half in these metrics; the model is one of the lowest in precision, recall, and AUC. The Big model (colored in purple) adds in data filtering to incorporate pseudo-masks with an area greater than 1\% into the next iteration of self-training, and it improves the performance from the vanilla model, supporting data filtering and the intuition that we tend to be more confident with larger masks. The Big, Convex model (colored in red) additionally utilizes the convex hull and achieves one of the best performances for all metrics. The Big, Convex, Fill model (colored in green) incorporates hole filling and achieves consistently better results, especially in recall and AUC. Adding the logic to not replace with empty pseudo-masks, the model (colored in yellow) performs better than most models in accuracy, precision, and AUC. When the self-training alternates between using large masks and large, connected masks (colored in blue), the performance drops to middle tier. This is likely because the pseudo-masks left out represents a significant portion, leaving the training set still dominated by the source domain with a heavy domain shift. 

Combining large masks, the technique of convex hull, and hole filling resulted in the best performance in iteration 2 with an accuracy of 0.7563, a precision of 0.6042, a recall of 0.7436, and an AUC of 0.8220. However, a limitation of this study is the lack of a termination rule to stop the self-training, and our future efforts will be dedicated to addressing this aspect. Another constraint is the relatively small dataset in the source domain, which is smaller than our target domain.

\begin{figure*}
  \centering
  \begin{subfigure}{0.115\linewidth}
    \includegraphics[width=\linewidth]{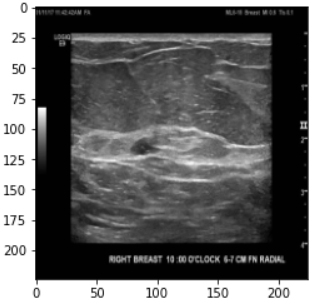}
  \end{subfigure}
  \hfill
  \begin{subfigure}{0.115\linewidth}
    \includegraphics[width=\linewidth]{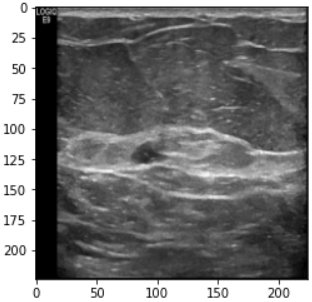}
  \end{subfigure}
    \hfill
  \begin{subfigure}{0.115\linewidth}
    \includegraphics[width=\linewidth]{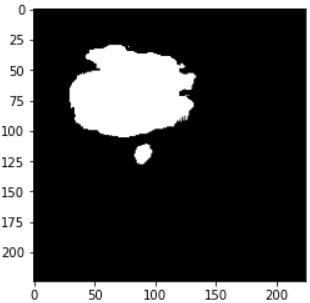}
  \end{subfigure}
  \hfill
  \begin{subfigure}{0.115\linewidth}
    \includegraphics[width=\linewidth]{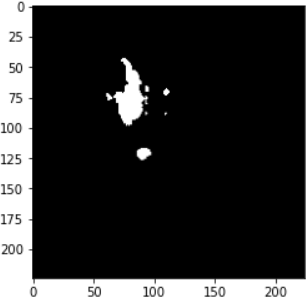}
  \end{subfigure}
  \hfill
  \begin{subfigure}{0.115\linewidth}
    \includegraphics[width=\linewidth]{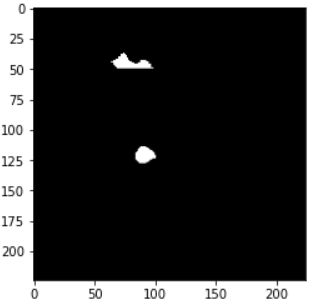}
  \end{subfigure}
  \hfill
  \begin{subfigure}{0.115\linewidth}
    \includegraphics[width=\linewidth]{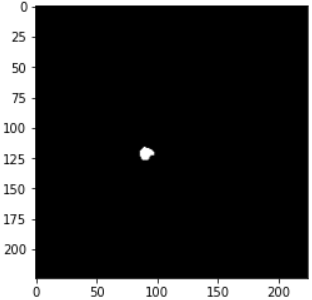}
  \end{subfigure}
  \hfill
  \begin{subfigure}{0.115\linewidth}
    \includegraphics[width=\linewidth]{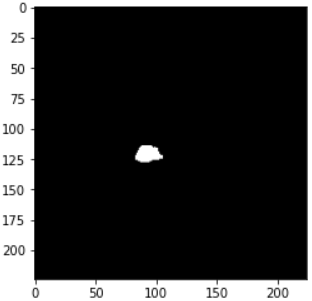}
  \end{subfigure}
  \hfill
  \begin{subfigure}{0.115\linewidth}
    \includegraphics[width=\linewidth]{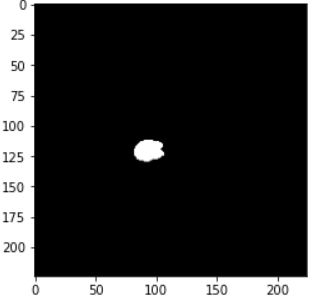}
  \end{subfigure}

  \begin{subfigure}{0.115\linewidth}
    \includegraphics[width=\linewidth]{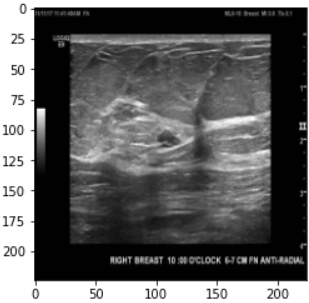}
  \end{subfigure}
  \hfill
  \begin{subfigure}{0.115\linewidth}
    \includegraphics[width=\linewidth]{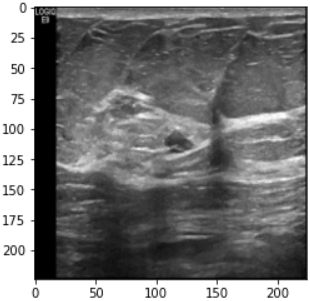}
  \end{subfigure}
    \hfill
  \begin{subfigure}{0.115\linewidth}
    \includegraphics[width=\linewidth]{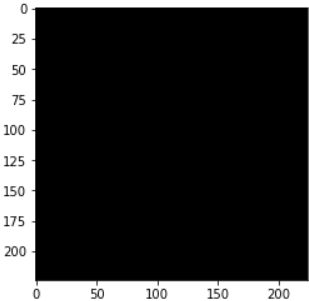}
  \end{subfigure}
  \hfill
  \begin{subfigure}{0.115\linewidth}
    \includegraphics[width=\linewidth]{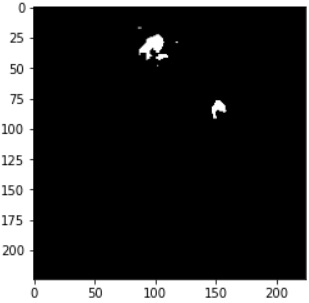}
  \end{subfigure}
  \hfill
  \begin{subfigure}{0.115\linewidth}
    \includegraphics[width=\linewidth]{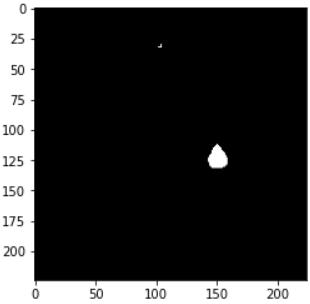}
  \end{subfigure}
  \hfill
  \begin{subfigure}{0.115\linewidth}
    \includegraphics[width=\linewidth]{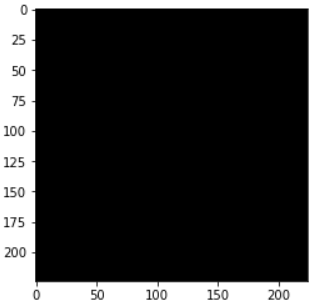}
  \end{subfigure}
  \hfill
  \begin{subfigure}{0.115\linewidth}
    \includegraphics[width=\linewidth]{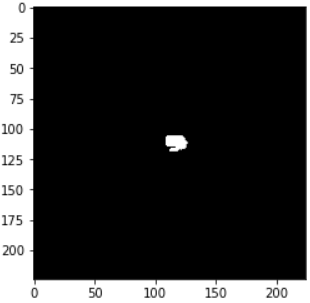}
  \end{subfigure}
  \hfill
  \begin{subfigure}{0.115\linewidth}
    \includegraphics[width=\linewidth]{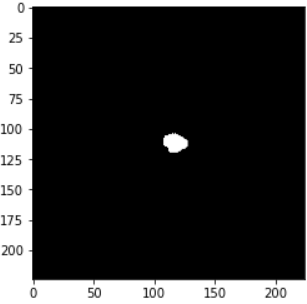}
  \end{subfigure}

  \begin{subfigure}{0.115\linewidth}
    \includegraphics[width=\linewidth]{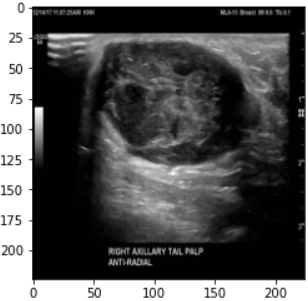}
    \caption{}
  \end{subfigure}
  \hfill
  \begin{subfigure}{0.115\linewidth}
    \includegraphics[width=\linewidth]{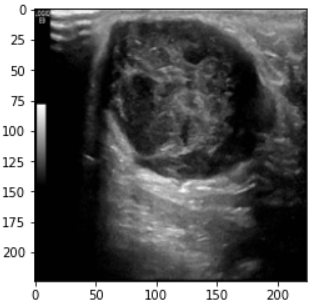}
    \caption{}
  \end{subfigure}
    \hfill
  \begin{subfigure}{0.115\linewidth}
    \includegraphics[width=\linewidth]{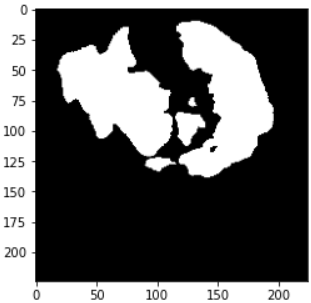}
    \caption{}
  \end{subfigure}
  \hfill
  \begin{subfigure}{0.115\linewidth}
    \includegraphics[width=\linewidth]{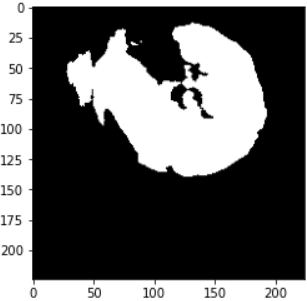}
    \caption{}
  \end{subfigure}
  \hfill
  \begin{subfigure}{0.115\linewidth}
    \includegraphics[width=\linewidth]{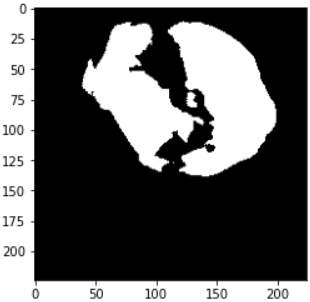}
    \caption{}
  \end{subfigure}
  \hfill
  \begin{subfigure}{0.115\linewidth}
    \includegraphics[width=\linewidth]{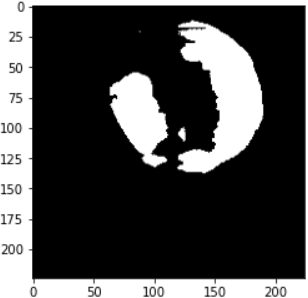}
    \caption{}
  \end{subfigure}
  \hfill
  \begin{subfigure}{0.115\linewidth}
    \includegraphics[width=\linewidth]{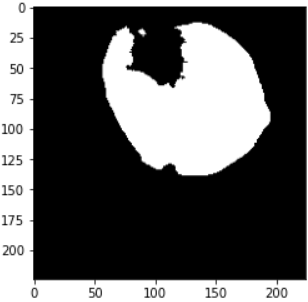}
    \caption{}
  \end{subfigure}
  \hfill
  \begin{subfigure}{0.115\linewidth}
    \includegraphics[width=\linewidth]{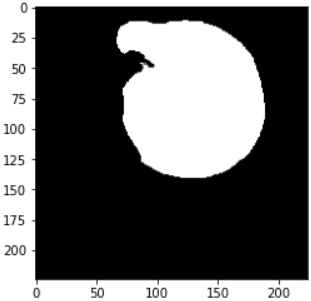}
    \caption{}
  \end{subfigure}
  \caption{Pseudo-mask generation on 3 target images (rows). (a) Original Image (b) Cropped Image (c) Pseudo-mask generated by the initial Teacher model (d)-(h) Pseudo-masks generated through iterations of self-training.}
  \label{fig:results}
\end{figure*}

Pseudo-masks for 3 images generated by our proposed approach are shown in Figure \ref{fig:results}. The initial Teacher model is trained solely on the source dataset and used to generate the initial pseudo-masks presented in Fig.~\ref{fig:results}(c). Compared to Fig.~\ref{fig:results}(h) which has the visually more accurate masks (generated in the fifth iteration of self-training), we observe the domain shift between two datasets though they are both sonographic breast images. Especially in the second row, the initial pseudo-mask is entirely blank. Through five rounds of self-training, a connected cluster of pixels is activated in the middle, corresponding to the shaded region in the middle of the mass. Though the pseudo-mask is predicted blank in the third iteration (Fig.~\ref{fig:results}(f)), the ROI emerges in the next iteration and even better than previous time (Fig.~\ref{fig:results}(c)) when it was also a blank mask. The first row has a similar process but with a non-empty pseudo-mask throughout all iterations, same as most images in the target domain. The third row had a better initial pseudo-mask and through self-training, it becomes complete and compact. These results show that the pseudo-masks converge in a few iterations.

\begin{figure}
  \centering
   \includegraphics[width=0.5\linewidth]{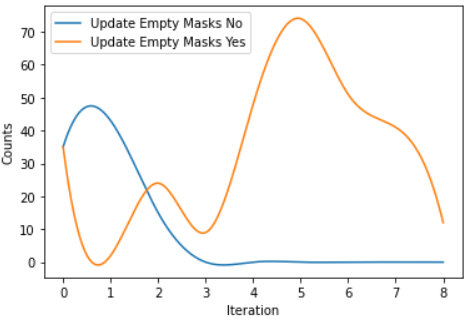}
   \caption{Number of images with empty pseudo-masks. Blue and orange lines represents with and without the additional logic on mask update, respectively.}
   \label{fig:maskupdate}
\end{figure}

\section{DISCUSSION}
One advantage is that through our proposed algorithm we can overcome the domain shift caused by unknown discrepancies such as different imaging systems, modalities, or styles. The approach is also suitable when there is no domain shift, so the adoption of this algorithm requires no check or quantifying the domain shift between two datasets on the same topic, as long as the images are pre-processed to be visually similar. When our approach is applied on raw images (see Figure~\ref{fig:entropy}(a)), the majority of pseudo-masks are blank or out of place due to a heavy domain shift, but after pre-processing with the entropy filter, our approach is able to iteratively refine the masks. In a way this represents a limitation of our algorithm that pre-processing can be required, but on the other hand, this can be seen as an advantage that preliminary pre-processing to make the images look alike is sufficient for the pre-condition of this algorithm. 

In cases where the pseudo-masks are blank, we default to use the entire cropped image for classification. With the additional update rule to not replace the previous pseudo-masks with new empty masks, we are bound to have more non-empty masks. The reasoning behind it is that blank masks provide no information to work with in the next iteration of self-training due to the fact that there must be a ROI in the image, so as long as we keep some information in play, utilizing the better pseudo-masks in training will effectively lead us to convergence. Figure \ref{fig:maskupdate} tracks the number of images with empty pseudo-masks throughout the iterations of self-training. Blue line shows the number of empty pseudo-masks rapidly decreasing to zero when the update rule is applied so that previous pseudo-masks are not replaced by empty ones. Orange line shows fluctuating amount of empty pseudo-masks when the update rule is not applied. Though a non-empty mask does not validate its correctness, this practice prevents good pseudo-maps to be overwritten by empty masks.

\section{CONCLUSION}
We presented an automated approach for producing image masks for the elimination of lesion ROI, specifically tailored for US breast imaging and differential diagnosis of benign and malignant breast masses. Through autonomous generation and iterative refinement of ROI masks via segmentation, using a U-Net architecture, our method achieved superior breast cancer classification results, providing a practical solution for various applications with sparse annotated data. The iterative self-learning process, using a preliminary model trained on a small open-source US dataset with ground truth, successfully produced pseudo-masks for a substantial unannotated private dataset without mask annotations. The efficacy of this pseudo-mask generation, guided and assessed by its performance in the downstream classification task facilitated by DenseNet169, highlights the potential of our approach in reshaping the landscape of AI-assisted medical image analysis and image-based disease diagnosis. This preliminary study lays the foundation for a paradigm shift in the field, prioritizing adaptability, efficiency, and enhanced diagnostic outcomes.

By assimilating insights from available datasets and extending our framework to diverse medical applications, our approach not only addresses the current challenges in US breast imaging but also opens avenues for more effective and efficient generalization in leveraging deep learning for enhanced healthcare outcomes. In our future work, we plan to refine our methodology to enhance the convergence of classification performance. Furthermore, our focus extends to establishing a rule for terminating the self-training process. These efforts are critical for improving the overall effectiveness of our approach. Through the implementation of refined convergence strategies and a well-defined termination criterion, we aim to strengthen the algorithm, ensuring superior performance and versatility in practical applications.


\end{document}